\documentclass[11pt]{article}

\usepackage{ACL2023}

\usepackage[english]{babel}
\usepackage[nottoc]{tocbibind}
\usepackage{times}
\usepackage{latexsym}
\usepackage{cuted}
\usepackage{amsfonts}
\usepackage{tabularx}
\usepackage{booktabs}
\usepackage{multirow}
\usepackage{graphicx}
\usepackage{amsmath}
\usepackage{diagbox}
\usepackage{multicol}
\usepackage{mathrsfs}
\usepackage{algorithm}
\usepackage{algpseudocode}
\usepackage{listings}
\usepackage{color}
\usepackage[T1]{fontenc}
\usepackage{subfigure}
\usepackage[utf8]{inputenc}
\usepackage{microtype}
\usepackage{inconsolata}
\newcommand{\name}{\textit{Attention Buckets}}

\title{Fortify the Shortest Stave in Attention: Enhancing Context Awareness of Large Language Models for Effective Tool-Use}

\author{
\begin{tabular}{c}
Yuhan Chen$^{1}$\thanks{\ \ Equal contribution.} \quad \quad  Ang\  Lv$^{1}$\footnotemark[1] \\\textbf{Ting-En Lin}$^{2}$ \quad \textbf{Changyu Chen}$^{1}$ \quad \textbf{Yuchuan Wu}$^{2}$\\ 
\textbf{Fei \  Huang}$^{2}$ \ \quad \quad \textbf{Yongbin Li}$^{2}$\footnotemark[2] \quad \quad \ \ \textbf{Rui \ Yan}$^{1,3}$\thanks{\ \ Corresponding authors.} 
\end{tabular}
\\ \vspace{.5mm}
    \small
    \begin{tabular}{c}
    $^1$Gaoling School of Artificial Intelligence, Renmin University of China \quad $^2$Alibaba Group\\
    $^3$Engineering Research Center of Next-Generation Intelligent Search and Recommendation, Ministry of Education\\
    \end{tabular}
    \\ \vspace{.5mm}
    \small
    \begin{tabular}{c}
    \texttt{\{yuhanchen, anglv, chen.changyu, ruiyan\}@ruc.edu.cn} \\
    \texttt{\{ting-en.lte, shengxiu.wyc, f.huang, shuide.lyb\}@alibaba-inc.com}\\
    \end{tabular}
    \vspace{2mm} \\
}

\begin{document}
\maketitle
\begin{abstract}
In this paper, we demonstrate that an inherent waveform pattern in the attention allocation of large language models (LLMs) significantly affects their performance in tasks demanding a high degree of context awareness, such as utilizing LLMs for tool-use.
Specifically, the crucial information in the context will be potentially overlooked by model when it is positioned in the trough zone of the attention waveform, leading to decreased performance.
To address this issue, we propose a novel inference method named \name.
It allows LLMs to process their input through multiple parallel processes. 
Each process utilizes a distinct base angle for the rotary position embedding, thereby creating a unique attention waveform.
By compensating an attention trough of a particular process with an attention peak of another process, our approach enhances LLM's awareness to various contextual positions, thus mitigating the risk of overlooking crucial information.
In the largest tool-use benchmark, our method elevates a 7B model to achieve state-of-the-art performance, comparable to that of GPT-4.
On other benchmarks and some RAG tasks, which also demand a thorough understanding of contextual content, our \name\ also exhibited notable enhancements in performance.\footnote{We release our code at \url{https://github.com/Fiorina1212/Attention-buckets}.}
\end{abstract}

\vspace{2mm}
\section{Introduction}
\label{sec:intro}
Recent works that augmenting large language models (LLMs, e.g., GPT series~\cite{brown2020language,openaichatgptblog,openai2023gpt4}) with tools have achieved advancements in various fields, such as human-computer interactions~\cite{qin2023toolllm,schick2023toolformer}, automating multi-modal tasks~\cite{surís2023vipergpt,patil2023gorilla}, and enhancing the overall efficiency of language-related applications~\cite{shen2023hugginggpt}.
In this paradigm, upon receiving a user's intent, a large language model accesses multiple tools, typically in the form of APIs. 
It then selects the most suitable one by referring to the relevant tool documentation, and provides an accurate and suitable response.
Considering the integration of extensive information into the context, tool-use tasks demand a high level of context understanding and awareness from LLMs.

\begin{figure*}
\vspace{2mm}
    \centering
    \includegraphics[width=\linewidth]{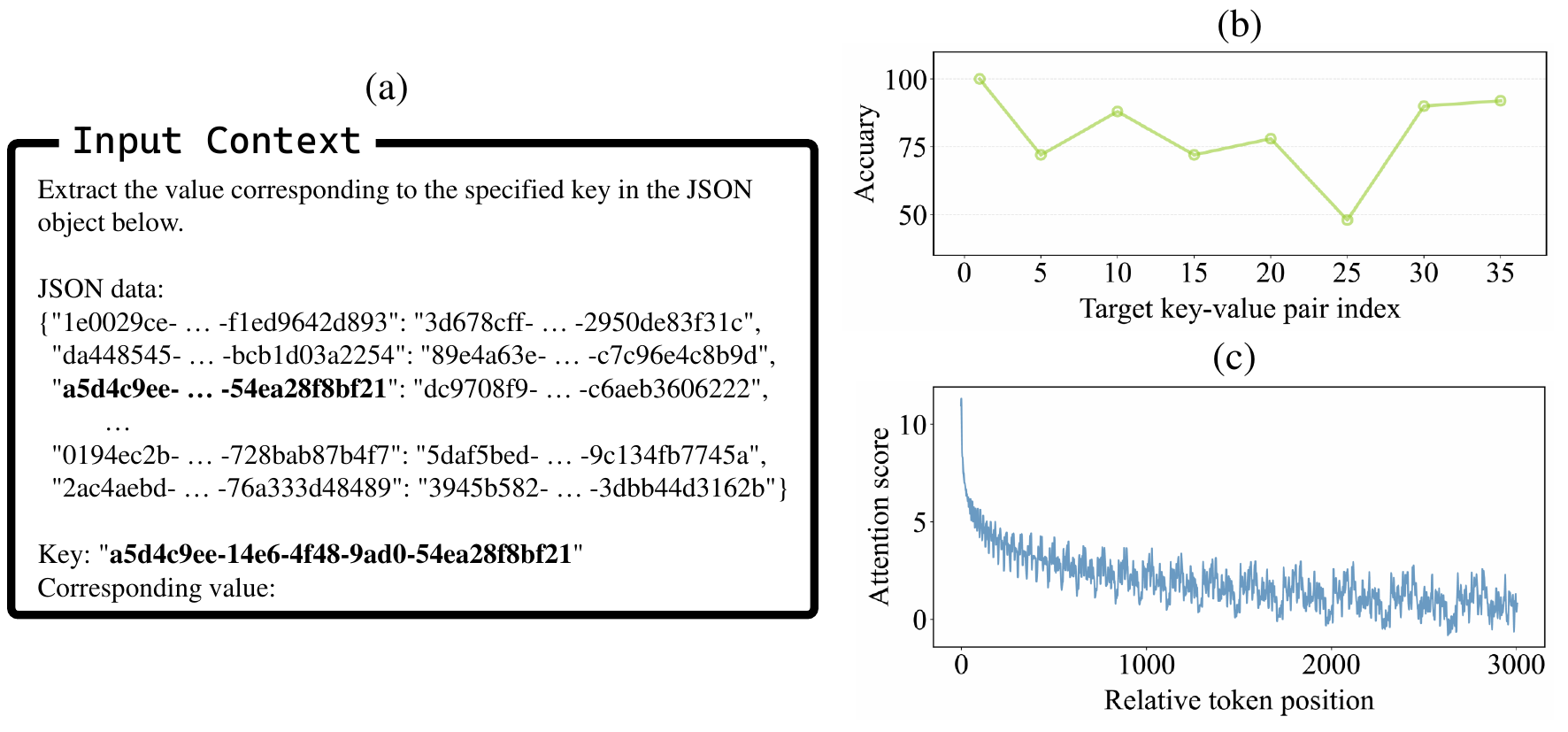}
    \caption{
    (a) Task illustration: Presented with multiple key-value pairs and a target key (highlighted in bold), the model is required to accurately retrieve and generate the value associated with this key from an extensive context.
    (b) We illustrate the position-related fluctuation in accuracy of Llama-2-7B on this in-context retrieval task.
    (c) The pattern of the attention score exhibits fluctuations, which we term the ``attention waveform''. 
    Our study reveals a connection between the position-related fluctuations in LLMs' performance and this attention waveform.}
    \label{fig:motivation}
\end{figure*}

Despite the achievements made by current LLM-based tool-use frameworks, in our practical experience, we observed that LLMs exhibit varying levels of awareness concerning different positions within the context.
For instance, LLMs may overlook certain tools within the context, resulting in a failed call; however, by altering the position of these tools, the task can be successfully executed.
Such variations significantly affect the performance of LLMs in tool-use.
This observation is consistent with the findings from a previous study~\cite{liu2023lost} that investigated a simple in-context retrieval task.
When LLMs are presented with multiple key-value pairs and instructed to retrieve the value associated with a specific key, the index of the queried target key results in significant fluctuations in accuracy. 
Figure~\ref{fig:motivation}(a) provides a visual representation of the instructions for this task. 
Figure~\ref{fig:motivation}(b) shows this fluctuation we replicated using the Llama-2-7B~\cite{touvron2023llama2}.
In our study, we go beyond the superficial fluctuations previously observed and identify that these position-related performance differences are closely associated with the model's fluctuating attention allocation.
Specifically, we observed a waveform pattern in the attention ``intensity'' (referred to as the attention waveform in this paper) when LLMs retrieve the same token from the context, as illustrated in Figure~\ref{fig:motivation}(c). 
We demonstrate that if the position of the crucial information coincides with a trough in the attention waveform, the model may overlook it, leading to decreased accuracy. 

Based on insight above, we argue that by shifting essential information away from the attention waveform's trough zone, we can reduce the risk of LLMs\footnote{In this paper, we focus on LLMs based on Transformer models~\cite{NIPS2017_3f5ee243} and rotary position embeddings (RoPE~\cite{su2022roformer}). 
This family of LLMs include many popular models like Llama~\cite{touvron2023llama1,touvron2023llama2}, Qwen~\cite{bai2023qwen}, Baichuan~\cite{yang2023baichuan}, etc.
} missing crucial details, thus enhancing the efficacy of tool-use.
Because crucial information within the context is inaccessible in practice, we propose the following approach to circumvent this challenge:
We process the context through multiple parallel executions, where each execution is assigned a unique rotary angle base of the rotary position embedding, resulting a distinct waveform pattern (See \S\ref{sec:rope} for details).
By ensuring these attention waveforms are ``complementary,'' — for any position where one waveform reaches its trough, another waveform reaches its peak — we enhance the LLM's context awareness across various positions.
We then aggregate the output distributions from these parallel executions and compute their weighted sum. 
This sum is subsequently decoded to generate the final prediction token.

An analogy can aid in understanding our approach: Imagine a wooden bucket with some shorter staves, which allow water to leak out. 
Similarly, the attention mechanism, at each angle base, has limited awareness of specific positions in the context. 
We utilize models to process the context with different angle bases. 
This results in the troughs of one attention wave being fortified by the peaks of another, analogous to how the longer staves in one bucket compensate for the shorter staves in another. 
Consequently, we name our proposed method \name.

We achieve the state-of-the-art on the largest tool-use benchmark ToolBench~\cite{qin2023toolllm} and another benchmark ToolAlpaca~\cite{tang2023toolalpaca}.
In ToolBench, we augment the performance of a 7B LLM to levels competitive with those of GPT-4~\cite{openai2023gpt4}. 
In addition to our achievements in tool-use, we also demonstrate our method's potential in general retrieval-augmented generation (RAG) tasks, which also demand a high degree of contextual awareness.
In summary, we make three major contributions:

(1) For LLMs with RoPE, we propose and verify an explanation for the variation in their awareness of different positions within the context. 
We establish a relationship between this variation and the attention waveform.

(2) By leveraging the insights from our proposed explanation, we develop a novel approach \name\ to enhance LLMs' context awareness.

(3) Through extensive experiments, we empirically validate the efficacy of our proposed method.


\section{Attention Waves Impact on Context Awareness}

In this section, we demonstrate that position-related performance fluctuations of LLMs are influenced by the underlying attention waveform.

\subsection{Preliminaries}
\label{sec:rope}
Rotary position embedding (RoPE)~\cite{su2022roformer} stands as a prevalent technique for position encoding in large language models with Transformer backbone~\cite{NIPS2017_3f5ee243}.
During the attention calculation, given a query or key vector at position $m$ in the sequence, RoPE serves to encode the position information into the vector via a $d$-dimensional rotation matrix denoted as $R_{\theta,m}$. 
This matrix $R_{\theta,m}$ is structured as a block diagonal matrix consisting of blocks with dimensions of $2\times2$, totaling $d/2$ such blocks. 
Specifically, the $i$-th block is defined as:
\begin{equation}
    \begin{aligned}
    R_{\theta_i, m} = \begin{bmatrix}
         \cos m \theta_{i} & -\sin m \theta_{i}\\
         \sin m \theta_{i} & \cos m \theta_{i}\\
    \end{bmatrix},
    \end{aligned}
\end{equation}
where $\theta_i = B^{-\frac{2i}{d}}$, with $B$ is termed as the base of the rotary angle. 

In each Transformer layer, after multiplying the query vector $q_{m}$ at position $m$ and the key vector $k_{n}$ at position $n$ with the rotation matrix, the relative position is incorporated in their inner product (the attention score before softmax):
\begin{equation}
    \begin{aligned}
        (R_{\theta,m} q_{m})^{\top} (R_{\theta,n} k_{n}) = q^{\top}_{m} R_{\theta, n-m} k_{n}.
    \end{aligned}
\end{equation}

When the relative distance $n-m$ increases, the waveform of the attention score before softmax demonstrates a long-term decay, i.e., the value generally decreases as the relative distance grows. 
This trend is accompanied by a waveform, as depicted in Figure~\ref{fig:motivation}(c).
The derivation of this waveform is presented in Appendix~\ref{apx:upper-bound}.

As a widely-used position embedding technique in LLMs, many researchers found RoPE has a substantial impact in LLM's context utilization and awareness.
RoPE has favorable properties that enhance the model in various aspects, such extrapolation~\cite{chen2023extending,gao2023empower,liu2023scaling}, efficient long-context model training~\cite{zhu2023pose,xiong2023effective}, and better understanding of training data~\cite{lv2023falling}.
In this paper, we leverage the attention waveform introduced by position embeddings to enhance the context awareness of LLMs.
We hypothesize that these waveform patterns might affect the model's context awareness. 
Intuitively, tokens located at troughs of the attention waveform would receive less focus. 
If such tokens are important for the current prediction, this could hamper the performance.
We designed an experiment to test this hypothesis.

\subsection{Hypothesis Verification}
\label{sec:verify}
\paragraph{Task and Data} 
We conducted an in-context retrieval test~\cite{liu2023lost,longchat2023}.
We feed the Llama-2-chat-7B~\cite{touvron2023llama2} with $K$ synthetic key-value pairs in JSON format. 
Each key and value is a distinct UUID string~\cite{rfc4122}.
We then prompt the model to retrieve the value corresponding to the key we specify.
We evaluated the model's context awareness based on the accuracy of the value it generates. 
Figure~\ref{fig:motivation}(a) shows a test example.

\paragraph{Experiment Design} 
We varied the RoPE base within the model from $10,000$ to $30,000$ in increments of $5,000$. 
For each base value, we calculate the corresponding waveform of attention score and identify the positions of the peaks and troughs (see Appendix~\ref{apx:calcu} for details).
Each test sample undergoes two evaluation rounds:
In the first round, we position the target key-value pair at the attention peak nearest to the exact middle of the context.
In the second round, we move the target pair to the nearest attention trough. 
By comparing accuracy differences between the two rounds, we aimed to answer how much attention waveform patterns impact the model's context awareness. 
The experiment was conducted with varying context lengths by setting different $K$ (40 and 50, respectively).
We provide more experimental details in appendix~\ref{apx:kv}.

\begin{table}[t]
    \centering
    \resizebox{0.96\linewidth}{15mm}{
    \begin{tabular}{lcccc}\toprule
    \multirow{2}{*}{Base} & \multicolumn{2}{c}{$K$=40} & \multicolumn{2}{c}{$K$=50} \\
    \cmidrule(l{2pt}r{2pt}){2-3}\cmidrule(l{2pt}r{2pt}){4-5} & Peak Acc & Trough Acc & Peak Acc & Trough Acc\\\hline
        10,000  & 79.8 & 76.8 & 47.6 & 44.0\\
        15,000  & 96.6 & 96.2 & 75.8 & 75.2 \\
        20,000  & 85.2 & 85.0 & 82.6 & 80.4\\
        25,000  & 70.8 & 70.0 & 59.2 & 55.6\\
        30,000  & 62.2 & 57.6 & 51.8 & 24.4\\
        \bottomrule
    \end{tabular}}
    \caption{The results of the in-context key-value retrieval.
    The generation accuracy provides insight into the model's awareness of information at both the peaks and troughs of the attention waveform.}
    \label{tab:fluc}
\end{table}

\begin{figure*}
    \centering
    \includegraphics[width=0.96\linewidth]{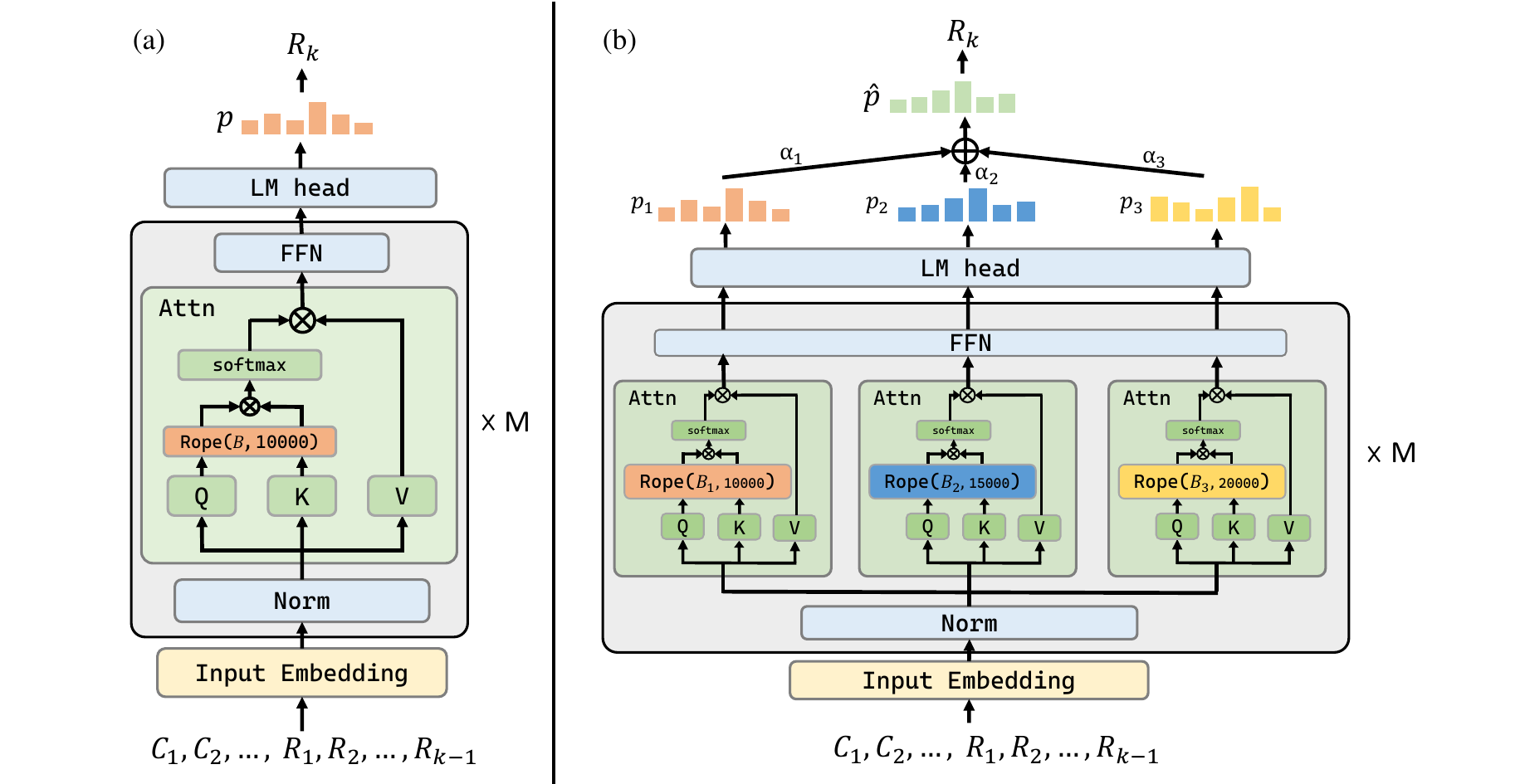}
    \caption{(a) The overview of how a typical Transformer-based Large Language Model (LLM) generates the next token based on context \(\mathcal{C}\). This LLM comprises \(M\) layers, though for simplicity, only the inner workings of a single layer are shown.
    (b) The overview of our proposed \name\ augmenting the context awareness of LLMs: Upon receiving context \(\mathcal{C}\), it creates \(N\) (specifically 3 in this example) parallel copies for processing. 
    Each parallel stream employs a distinct RoPE base. 
    The resulting output distributions \(p_j\) are weighted and summed based on the prediction confidence \(\alpha_j\), culminating in the final predicted distribution \(\hat{p}\) used for decoding the next token.}
    \label{fig:overview}
\end{figure*}

\subsection{Results and Analysis}
\label{sec:2.3}
Our experimental findings, as detailed in Table~\ref{tab:fluc}, reveal a performance trend associated with varying RoPE base values: we observe an initial rise followed by a subsequent fall.
Notably, placing a key-value pair at the peak of the attention waveform consistently yields better outcomes than positioning it at the trough. 
This holds true across different context lengths (as defined by $K$) and base values.
Also, the results suggest that the optimal base values differ depending on the context length. 
For example, with $K$ set at 40, the best performance is achieved with a base of 15,000, while a base of 20,000 is most effective when $K$ is 50.

Based on these results, we can draw the following insight and its associated challenge:

\textbf{\textit{Insight}}: Enhancing the attention to information positioned at the troughs of the attention waveform could make the context awareness of large language models more robust, potentially leading to improved overall performance.

\textbf{\textit{Challenge}}: In practical applications, pinpointing the location of critical information is difficult. 
This makes it challenging to select a RoPE base that ensures attention to the crucial information.

\section{Enhancing Context Awareness via Interleaving Attention Waveform}

Based on the insight presented above, we introduce a novel approach to sidestepping the above challenge, with the goal of improving the LLM performance in tool-use by enhancing its context awareness.
Our method focuses on the inference stage of LLMs and does not require training. 
We first provide preliminary definitions, followed by a detailed introduction to our approach.

\subsection{Preliminaries}

In tool-use, fulfilling a user's intent typically involves multiple turns, such as selecting tools, calling for APIs, and engaging with the user across multiple interactions. 
In this section, our introduction focuses on one single turn since the multi-turn scenario is a simple amalgamation of this single-turn scenario.
We consider all information from previous turns, including API responses, tool execution outcomes, and user feedback, as the model's context for the current turn, which we represent as $\mathcal{C}$. 
Subsequently, the LLM proceeds to generate a response, denoted as $\mathcal{R}$, in an autoregressive manner based on $\mathcal{C}$. 
To denote the specific tokens within $\mathcal{C}$ or $\mathcal{R}$, we employ the notation $\mathcal{C}_{k}$ or $\mathcal{R}_{k}$, respectively, where $k$ represents the token's index.

\subsection{Method}
Given an input context $\mathcal{C}$, our approach involves duplicating this context into $N$ copies, forming a batch that allows for parallel processing by the LLM. 
In each parallel, each of these $N$ copies is individually processed with a distinct RoPE base $B_j$ from a base set \( \mathcal{B}_{c} \), resulting in $N$ corresponding predicted distribution $p$ over the vocabulary $V$.
Our selection of $\mathcal{B}_{c}$ guarantees that an attention trough in one parallel is compensated by a peak in another, effectively reducing the possibility of the LLM missing essential information residing within an attention trough. 
We will delve into the details of determining $\mathcal{B}_{c}$ in $\S$~\ref{sec:search}.

We posit that in the parallel run indexed by $j$, if the model focuses its attention on crucial information it currently requires, it has more confidence to make accurate predictions for the next token in the response $\mathcal{R}$. 
We quantify the model's confidence on prediction $\alpha_j$ as:
\begin{equation}
\begin{aligned}
    &\alpha^{'}_j = \max_{v \in V} \ p(\mathcal{R}_{k} = v | \mathcal{C}, B_j, \mathcal{R}_{1:k-1}),\\
    &\alpha_j = \frac{e^{\alpha^{'}_j}}{\sum_{i=1}^{N} e^{\alpha^{'}_i}}.
\end{aligned}
\end{equation}

Next, we compute a weighted sum of each run's output distribution $p_j$ to derive the final predicted distribution $\hat{p}$. 
The weighting of each $p_j$ depends on its corresponding confidence score $\alpha_j$:
\begin{equation}
\hat{p} = \sum^{n}_{j} \alpha_j * p_j.
\end{equation}

We decode a predicted token from $\hat{p}$. 
This token is incorporated into the preceding context, and this auto-regressive process persists until the current turn ends.

\subsection{The Searching of $\mathcal{B}_{c}$}
\label{sec:search}
This section details our methodology for searching $\mathcal{B}_{c}$, an appropriate set of RoPE bases. 
Our goal is to develop strategies ensuring that the attention waveform troughs of any given base overlap with peaks from different bases, and vice versa. 
Firstly, we define a discrete base search space, denoted as:
\begin{equation}
\resizebox{0.96\linewidth}{!}{
$\mathcal{B}_{s} = \left\{B_{i} \middle| B_{i} = B_{\text{min}} + i \times S, \ i \in \left(0, \frac{B_{\text{max}}-B_{\text{min}}}{S}\right] \right\},$}
\label{eq:search-space}
\end{equation}
where \( B_{\text{min}} \) and \( B_{\text{max}} \) represent the minimum and maximum base values, and \( S \) is the search stride. 
In our experimental setup, we set \( B_{\text{min}} \) equal to \( B_{\text{train}} \), the base used during model pre-training.
This decision is grounded in the consideration~\cite{liu2023scaling} that opting for a smaller base compared to the one used during pre-training could potentially introduce out-of-distribution (OOD) positional information, as discussed in detail in the appendix~\ref{apx:ood-pos}.

At the beginning of the search, we initialize \( \mathcal{B}_{c} \) to \( \{B_{\text{train}}\} \).
For the following $N-1$ iterations, we search for a candidate base value in each round to be included in $\mathcal{B}_{s}$. 
In every round, we first identify the peaks and troughs within the waveform associated with each base in $\mathcal{B}_{s}$ and $\mathcal{B}_{c}$.
The selection of a candidate is determined by measuring the distance between the position of the $i$-th peak (and trough) for a candidate base and that of the $i$-th trough (and peak) for bases within set $\mathcal{B}_{c}$.
The maximum position of peaks or troughs that we take into account is constrained by the maximum context length.
The candidate with the shortest average distance is subsequently included in $\mathcal{B}_{c}$.
Our searching algorithm is detailed in Algorithm~\ref{alg:search}.

\begin{figure*}[t]
\vspace{2mm}
    \centering
    \includegraphics[width=0.92\linewidth]{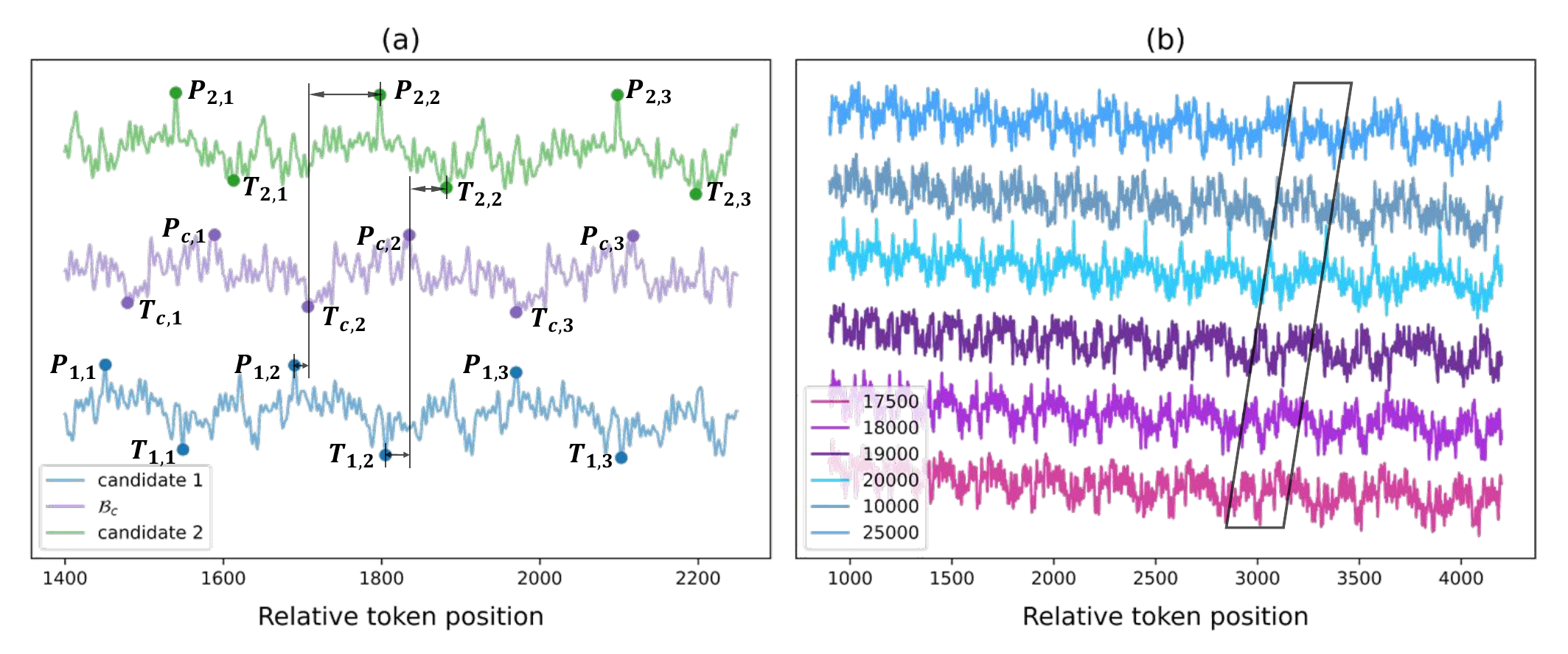}
    \caption{(a) RoPE Base Searching: we measure the distance from the candidate bases' attention peaks (or troughs) to the attention troughs (or peaks) corresponding to items in $\mathcal{B}_{c}$. 
    For demonstration clarity, we illustrate only a partial context that corresponds to one waveform period.
    (b) Attention waveform corresponding to $\mathcal{B}_{c}$ searched by hyperparameters detailed in $\S$\ref{sec:search}.}
    \label{fig:selection}
\end{figure*}

Figure~\ref{fig:selection}(a) is an algorithm illustration, showcasing the initial round of the search where $\mathcal{B}_{c}$ consists of just one item, and there are only two candidates.
It is clear that $d_1 = \sum_{i=1}^{3} |P_{1,i} - T_{c,i}| + \sum_{i=1}^{3} |T_{1,i} - P_{c,i}| < d_2 = \sum_{i=1}^{3} |P_{2,i} - T_{c,i}| + \sum_{i=1}^{3} |T_{2,i} - P_{c,i}|$. 
Consequently, candidate 1 is chosen.

In Figure~\ref{fig:selection}(b), we demonstrate the searched $\mathcal{B}_{c}$ with the hyper-parameters \( B_{\text{min}} = B_{\text{train}} = 10,000 \), \( B_{\text{max}} = 30,000\), \( S = 500\), and \( N = 6\).
The values in our searched \( \mathcal{B}_{c} \) consist of $\{1.00$, $1.75$, $1.80$, $1.90$, $2.00$, $2.50\} \times 10^{4}$.
In this figure, we can sketch a parallelogram to help us observe the patterns of the waveforms. 
Each waveform features a peak point that can be positioned along the left edge of this parallelogram. These peak points effectively divide this edge into several equal segments. 
This suggests that our searched $\mathcal{B}_{c}$ possesses waveforms that are evenly and densely distributed, minimizing the likelihood of a position being overlooked.

\begin{algorithm}[t]
\caption{The searching algorithm of $\mathcal{B}_{c}$.}
\begin{algorithmic}[1]
\State \textbf{Input}:
\begin{itemize}
    \item $P_{c}$ and $T_{c}$: Sets containing the peak and trough positions in attention waveforms corresponding to items in $\mathcal{B}_{c}$. 
    These positions are calculated by functions $f_{p}$ and $f_{t}$ (see appendix~\ref{apx:calcu}), respectively.
    \item Searched set $\mathcal{B}_{c}$, initialized as $\{B_{\text{train}}\}$.
    \item Search space $\mathcal{B}_{s}$, initialized using Eq.~\ref{eq:search-space}.
\end{itemize}
\State $P_c \gets f_{p}(B_\text{train})$, $T_c \gets f_{t}(B_\text{train})$
\While{$|\mathcal{B}_{c}|$ < $N$}
\For{$B_j$ in $\mathcal{B}_{s}$}
    \State $P_j \gets f_{p}(B_j)$, $T_j \gets f_{t}(B_j)$
    \State $d_j \gets $\begin{small}$
    \sum\limits_{\substack{p_{j,i} \in P_{j}\\t_{c,i} \in T_c}}
  |p_{j,i} - t_{c,i}|+\sum\limits_{\substack{t_{j,i} \in T_j\\p_{c,i} \in P_c}} |t_{j,i} - p_{c,i}|$\end{small} 
\EndFor
\State $\mathcal{B}_{c} \gets \mathcal{B}_{c} \cup  \{B'_j \ \text{with the minimum} \ d_j \} $. 
\State $P_c \gets P_c \cup f_{p}(B'_j)$, $T_c \gets T_c \cup f_{t}(B'_j)$
\EndWhile
\State \textbf{Output}: $\mathcal{B}_{c}$.
\end{algorithmic}
\label{alg:search}
\end{algorithm}

\section{Experiments}

\subsection{Experiment Setups}
\paragraph{Benchmark}
Up to the time of this paper, Toolbench~\cite{qin2023toolllm} stands as the largest benchmark for evaluating the tool-use proficiency of large language models. 
It has extensive resources, including 3,451 tools, 16,464 APIs, 126,486 instances, and 469,585 API calls.
All api calls in Toolbench are real and sampled from Rapid API.

The Toolbench evaluation consists of three distinct levels and three specific scenarios, each offering its own set of challenges.
The three evaluation levels include \textbf{Inst.}: testing the model's response to new instructions for tools already covered in the training data; \textbf{Tool.}: measuring performance with unfamiliar tools within the same tool categories as those in the training dataset; and \textbf{Cat.}: examining the model's ability to handle tools from completely new categories not represented in the training data.
The scenarios are \textbf{I1}: single-tool instructions, \textbf{I2}: multi-tool instructions within the same category, and \textbf{I3}: multi-tool instructions spanning across different collections. 
Due to specific details, there are only six combinations of levels and scenarios: I1-Inst., I1-Tool., I1-Cat., I2-Inst., I2-Cat., and I3-Inst.
Each combination comprises 200 test queries, with the exception of I3-Institution, which includes 100 queries.
For a more detailed introduction, readers are recommended to ~\cite[$\S$3.2]{qin2023toolllm}.

\begin{table*}[t]
\vspace{3mm}
\centering
\resizebox{\textwidth}{30mm}{
\begin{tabular}{cccc|cc|cc|cc|cc|cc|cc}
 \toprule
\multirow{2}{*}{\textbf{Model}} & \multirow{2}{*}{\textbf{Method}}  & \multicolumn{2}{c}{I1-Inst.} & \multicolumn{2}{c}{I1-Tool.} & \multicolumn{2}{c}{I1-Cat.} & \multicolumn{2}{c}{I2-Inst.} & \multicolumn{2}{c}{I2-Cat.} & \multicolumn{2}{c}{I3-Inst.} & \multicolumn{2}{c}{Avg} \\
&  \multicolumn{1}{|c|}{} & pass & win & pass & win & pass & win & pass & win & pass & win & pass & win & pass  & win       \\ \hline
\multirow{2}{*}{ChatGPT}& \multicolumn{1}{|c|}{ReACT}   &   41.5 & -&  44.0 &  -  & 44.5 & - &  42.5  &  -   & 46.5  & -   & 22.0  &  -  & 40.2  & -          \\
     & \multicolumn{1}{|c|}{DFSDT}   &  54.5  & 60.5  & 65.0&   62.0& 60.5  & 57.3   & 75.0  & 72.0   & 71.5  & 64.8   & 62.0  & 69.0   & 64.8    & 64.3           \\\hline
\multirow{2}{*}{Claude-2}& \multicolumn{1}{|c|}{ReACT}   &  5.5  & 31.0  & 3.5   & 27.8  & 5.5   & 33.8  & 6.0   & 35.0  & 6.0   & 31.5  &  14.0  & 47.5    & 6.8& 34.4          \\
     & \multicolumn{1}{|c|}{DFSDT}   &  20.5  &38.0   & 31.0   & 44.3  & 18.5   & 43.3  &17.0    & 36.8  & 20.5   &33.5   & 28.0   & 65.0    &22.6 & 43.5          \\\hline
\multirow{2}{*}{Text-Davinci-003}& \multicolumn{1}{|c|}{ReACT}   & 12.0   &28.5   & 20.0    &35.3   &20.0    & 31.0  & 8.5   &29.8  &14.5&29.8 & 24.0   &45.0   &  16.5  & 33.2         \\
     & \multicolumn{1}{|c|}{DFSDT}   &  43.5  & 40.3  & 44.0   &43.8   & 46.0   & 46.8  & 37.0   & 40.5  & 42.0   &43.3   &46.0    &63.0     &43.1 & 46.3   \\\hline
\multirow{2}{*}{GPT4}& \multicolumn{1}{|c|}{ReACT}   & 53.5   &60.0  & 50.0    &58.8   &53.5    & 63.5  & 67.0   &65.8  &72.0&60.3 &47.0   &78.0   &  57.2  & 64.4         \\
     & \multicolumn{1}{|c|}{DFSDT}   &  60.0  & \textbf{67.5}  & \textbf{71.5}   &\textbf{67.8}   & \textbf{67.0}   & \underline{66.5}  & 79.5   & \underline{73.3}  & \textbf{77.5}   &63.3   &\textbf{71.0}    &\underline{84.0}    &\underline{71.1} & \underline{70.4}  
     \\\hline
\multirow{3}{*}{ToolLlama}& \multicolumn{1}{|c|}{ReACT}   &   25.0 & 45.0  & 29.0   & 42.0  & 33.0   & 47.5  &30.5    & 50.8  & 31.5   & 41.8  & 25.0   & 55.0    &29.0 &47.0           \\
     & \multicolumn{1}{|c|}{DFSDT}   &  57.0  & 55.0  &61.0    & 55.3  & 62.0   &54.5   & 77.0   & 68.5  & \underline{77.0}   & 58.0   & 66.0   &69.0     &66.7 & 60.0          \\
     & \multicolumn{1}{|c|}{DFSDT-Retriever} & 64.0 &  62.3 & 64.0   & 59.0   &60.5    &55.5   & \underline{81.5}  & 68.5  & 68.5   & 60.8  & 65.0   & 73.0    &67.3&63.1           \\\hline
& \multicolumn{1}{|c|}{ReACT}   &  31.5  & 45.0  &  32.0  &  42.5 &  33.5  &  49.0 &  31.5 & 65.0  & 32.0   & 42.0 &  28.0  & 58.0    & 31.3&   50.3     \\
     ToolLlama& \multicolumn{1}{|c|}{DFSDT}   &  \underline{66.5}  &    \textbf{ 67.5} & 61.5 & 62.0  &     62.0  &  65.5  &  78.0 &   71.5 & 73.0  & \textbf{66.5}&      67.0 &82.0 & 68.0 & 69.2          \\
     + \name & \multicolumn{1}{|c|}{DFSDT-Retriever} &\textbf{68.5} & \underline{ 65.0} & \underline{70.0}   & \underline{65.5}  & \underline{65.0} & \textbf{67.0} &\textbf{84.0 }  &  \textbf{ 78.0 }&  71.0   &  \underline{64.5}&  \underline{69.0} & \textbf{89.0}    &\textbf{71.3} &\textbf{71.5}    \\\bottomrule
\end{tabular}
}
\caption{The tool-use performance on ToolBench~\cite{qin2023toolllm}.
We highlight the leading results for each task with \textbf{bold fonts}, and denote the second-best performance with \underline{underlines}.
\name\ augment the ToolLlama with only 7B parameters to outperform GPT-4 in both overall pass rate and win rate.}
\label{tab:toolbench}
\vspace{2mm}
\end{table*}

\paragraph{Models and Evaluation}

Based on the training dataset in ToolBench, \cite{qin2023toolllm} fine-tuned a model named ToolLlama, building upon Llama-2-7B~\cite{touvron2023llama2}. 
The authors compare it with advanced close-source LLMs, including ChatGPT~\cite{openaichatgptblog}, Claude-2, Text-Davinci-003, and GPT-4~\cite{openai2023gpt4}.
We implement \name\ to enhance the performance of ToolLlama.
The $\mathcal{B}_{c}$ employed for this benchmark is the same as detailed in $\S$\ref{sec:search} and depicted in Figure~\ref{fig:selection}(b).
Following~\cite{qin2023toolllm}, we adopt multiple reasoning methods for each model, including ReACT~\cite{yao2023react}, DFSDT~\cite{qin2023toolllm, shinn2303reflexion}, and an API retriever~\cite{qin2023toolllm} augmentation for reducing noise in tool selection (DFSDT-Retriever).
We adopt the greedy decoding strategy.

Evaluation of these models is conducted using two metrics: pass rate and win rate. 
The pass rate accesses how many user queries are fulfilled. 
The win rate, determined by ChatGPT evaluates the superiority of the model's solutions compared to those provided by ChatGPT-ReACT.

\subsection{Results and Analysis}
We present our experimental findings in Table~\ref{tab:toolbench}. 
Our \name\ enhances the scores of ToolLlama in almost every task level and scenario.
Notably, when paired with the DFSDT-Retriever setup (in the table's final row), our approach not only matches but often surpasses GPT-4's performance levels. 
On average, \name\ stand out, boasting the highest pass rate of 71.3$\%$ and win rate of 71.5$\%$. 
To our knowledge, \name\ set a new state-of-the-art (SOTA) result in this benchmark.

We also implement \name\ with various reasoning methods, as detailed in the table's bottom three rows. 
Each method showed marked improvements over their respective baselines, illustrating the versatility and compatibility of our approach. 
These results collectively indicate that \name\ boosts ToolLlama's tool-use proficiency, a success we attribute to its enhanced context awareness.

These accomplishments lead us to argue that language models harbor many untapped potentials. 
By effectively leveraging these capabilities, LLMs could be far more powerful than we thought. 
We hope our findings inspire further research into unlocking more fundamental abilities of LLMs.

We have also conducted additional experiments on other tool-use benchmarks, which are provided in appendix~\ref{apx:more-benchmarks} due to page limitations.

\subsection{Discussion on Efficacy}
\label{sec:efficacy}
Readers may have concerns about \name's efficiency, as parallel processing of context with varying base values could introduce additional memory overhead. 
However, it's important to note that all experiments described in this paper were successfully conducted using a single NVIDIA A100-80G GPU. 
Most importantly, \name\ does not compromise inference speed with sufficient memory.

To further address concerns regarding \name's effectiveness, we compare it with two methods:

$\bullet$ \textbf{$\name_{once}$.} Unlike the approach that utilizes $N$ inference processes with $N$ RoPE bases, $\name_{once}$ computes the average of $N$ attention waveforms from individual bases and then encodes the positional information using this averaged waveform. This technique utilizes only a single inference process, thereby avoiding any additional memory cost.

$\bullet$ \textbf{Attention Sorting.} In ASort~\cite{peysakhovich2023attention}, tokens in distant contexts that receive high attention are considered important. The authors first segmented the context and calculated the average attention of each segment. By rearranging segments in context based on sorted attention scores(with the highest attention segment placed last), they generate the answer using the newly sorted context. This approach does not require extra memory; however, it necessitates multiple iterations to gain the attention scores and lacks parallelizability.

$\bullet$ \textbf{Universal Self-Consistency.} USC~\cite{chen2023universal} is a universal self-consistency~\cite{wang2023selfconsistency} algorithm that supports free-format outputs. 
The LLM first generates $N$ responses.
Subsequently, the LLM is tasked with selecting the response that exhibits the highest degree of consistency, employing a specific prompt. 
The memory cost of this method is roughly equivalent to that of our \name.

\begin{table*}[t]
    \centering
    \resizebox{0.8\linewidth}{!}{
    \begin{tabular}{l|ll|ll|ll|ll|ll}\toprule
    \multirow{2}{*}{} & \multicolumn{2}{c|}{Original} & \multicolumn{2}{c|}{+$\text{AB}_{once}$} & \multicolumn{2}{c|}{+USC} & \multicolumn{2}{c|}{+ASort} & \multicolumn{2}{c}{+AB} \\
                      & pass      & win       & pass      & win       & pass        & win        & pass       & win    & pass       & win      \\ \hline
    I1-Inst.       & 64.0      &  62.3     & 52.0      & 37.0      & \underline{66.0}        & \underline{63.0}    &\underline{66.0}&   \underline{63.0} &\textbf{68.5}       & \textbf{65.0}       \\
    I1-Cat.        & 64.0      & 59.0      & 40.5      & 31.0      & 65.5      &61.5 &\underline{67.0}   & \underline{62.0}    & \textbf{70.0 }      & \textbf{65.5 }      \\
    I1-Tool.       &60.5       &55.5       & 47.0      & 34.5      & \underline{61.0}        & \underline{58.5}     & 59.5 &  \underline{58.5}  & \textbf{65.0 }      & \textbf{67.0}       \\
    I2-Inst.       &\underline{81.5}       & 68.5      & 70.5      & 65.0      & 80.0        & \underline{73.0}    & 80.0 & 68.5  & \textbf{84.0 }      & \textbf{78.0 }      \\
    I2-Cat.        & 68.5      & 60.8      & 65.0      & 58.0      & \underline{70.0 }       & \underline{61.5}   &68.5 & 60.3    & \textbf{71.0}       & \textbf{64.5}       \\
    I3-Inst.       & 65.0      & 73.0      & 61.0      & 52.0      & \underline{66.0}       & \underline{78.0}   & \underline{66.0} & 75.0  & \textbf{69.0}       & \textbf{89.0}       \\
    Avg            &67.3       &63.1       & 56.0      &45.3       & \underline{68.1}  &\underline{65.9}    &67.8&  64.6    & \textbf{71.3}       & \textbf{71.5}      \\
        \bottomrule
    \end{tabular}}
 \caption{Comparison among \name (AB), $\name_{once}$ ($\text{AB}_{once}$), Universal Self-Consistency (USC) and Attention Sorting (ASort), based on the ToolLlama-DFSDT-Retriever configuration. }
    \label{tab:ab1}
\end{table*}

All methods are evaluated using the ToolLlama-DFSDT-Retriever configuration.
The results are presented in Table~\ref{tab:ab1}.
Results reveal that preprocessing the aggregation of attention waveforms significantly reduces memory costs, but at the expense of the model's performance. 
Specifically, compared to the original \name, $\text{\name}_{once}$ shows a reduction of 11.3\% points in pass rate and 17.8\% points in win rate on average.
This decline is attributed to that the pre-averaged waveform can produce out-of-distribution position information. 
This issue does not arise in \name, where each base value is independently utilized during the forward computation.

ASort tries to strengthen the model's focus on key information through re-sorting, but its weak improvement in task performance, with a 0.5\% point increase in pass rate and 1.5\% point in win rate on average, reveals that the attention scores may not capture crucial information well. 

USC incurs a similar inference cost to our method. However, it only demonstrates a mere 0.8\% point increase in pass rate and a 2.8\% point increase in win rate on average, compared to the original ToolLlama. 
This limited enhancement can be attributed to the USC method's failure to effectively address the problem of trough position oversight inherent in inference with a single RoPE base. 
Despite multiple attempts, this oversight persists.
This comparison clearly illustrates the effectiveness of the \name\, as it outperforms the method with a similar level of overhead and without additional training.

\section{Exploring Applications for Retrieval-Augmented Generation}

\begin{table}[t]
    \centering
    \begin{tabular}{lccc}\toprule
    Method & NQ & WebQA \\\hline
        FiD-XL (3B) & 50.1 & 50.8 \\\hline
        Llama-2 (7B) & 48.5 & 51.7 \\
        + ASort&  48.9& 52.1 \\
        + USC & 47.6& 51.7 \\
         + \name & \textbf{50.3} {\tiny(1.8$\uparrow$)} & \textbf{53.1}{\tiny(1.4$\uparrow$)} \\
        \bottomrule
    \end{tabular}
    \caption{Accuracy on NQ and WebQ (10 documents).}
    \label{tab:qa}
\end{table} 

Considering our proposed \name\ enhances the model's context awareness, it should be effective in other tasks demanding high contextual information utilization. 
This section explores the effectiveness and generality of \name\ through a representative RAG task: open-domain question answering (ODQA).

Many current open-domain QA methods employ a Retrieval-Augmented Generation (RAG) paradigm~\cite{chen2017reading,khattab2021relevance, lewis2020retrieval}, where a retriever~\cite{qu2020rocketqa, khattab2020colbert} gathers relevant documents, followed by a generative reader~\cite{izacard2021leveraging, yu2022generate} to find answers, demanding high context awareness.

We conduct experiments on two popular benchmarks NaturalQuestion (NQ,~\cite{kwiatkowski2019natural}) and WebQA~\cite{berant2013semantic}, with 3,610 and 2,032 test samples, respectively. 
We assess the models based on their accuracy in providing answers.
An answer is considered correct if it contains one of the acceptable answers. 
In our experiments, we employed 10 documents as context and utilized the DPR~\cite{karpukhin2020dense} as the retriever, which is a supervised dense retrieval model trained on above datasets.

We use Llama-2-7B~\cite{touvron2023llama2} as our backbone model and compare it with FiD-XL~\cite{izacard2021leveraging}, the exclusive ODQA model trained on multiple ODQA benchmarks, including NQ and WebQ.

The results are shown in Table~\ref{tab:qa}. Compared with the other two methods, ASort~\cite{peysakhovich2023attention} and USC~\cite{chen2023universal}, \name\ exhibits a more stable improvement.
Additionally, when augmented with \name, Llama-2-7B demonstrated superior performance over the dedicated QA model FiD-XL.

\section{Ablation Study}
As the evaluation process for ODQA is both objective and convenient, it enables us to investigate whether our identified $\mathcal{B}_{c}$ approaches optimality through the examination of numerous permutations of base values.
Conducting this study on ToolBench is too costly for us, as it needs the prohibitive use of GPT-4 and ChatGPT for API calls.

We conducted an analysis of our search algorithm~\ref{alg:search} using the NQ dataset. 
We investigate the impact of varying \(N\), which represents the size of \(\mathcal{B}_{c}\), as well as \(S\), which corresponds to the search stride (i.e., granularity).
We generated four distinct variations of \(\mathcal{B}_{c}\).
The values of \(N\) and \(S\), along with their respective corresponding \(\mathcal{B}_{c}\) datasets, are provided in the upper section of Table~\ref{tab:ab}. 
Additionally, we compare these \(\mathcal{B}_{c}\) with the following variants:

1. Each individual element within \(\mathcal{B}_{c}\).

2. \(\mathcal{B}_{A.S.}\): This set is constructed with base values forming arithmetic sequences, having common differences of 3,000 and 4,000, respectively. Details of two \(\mathcal{B}_{A.S.}\) sets can be found in the lower section of Table~\ref{tab:ab}.

These comparisons enabled us to comprehensively evaluate the performance of our search algorithm concerning various search hyperparameters.
We present the results in Table~\ref{tab:ab2}, which reveal the following key conclusions:

Firstly, with various combinations of $N$ and $S$, \(\mathcal{B}_{c}\) searched by our algorithm~\ref{alg:search} consistently contribute to increased accuracy to a similar extent.
An arbitrary $\mathcal{B}_{c}$ also outperforms any $\mathcal{B}_{A.S.}$.
There results highlight the robustness and effectiveness of our method.

Most importantly, the enhancement of our method brought to LLMs aligns with our expectations, showing that various bases contribute to context awareness at different positions, rather than being reliant on specific ``optimal'' base values.
Note that $\mathcal{B}_{c4}$ has only one additional element compared to $\mathcal{B}_{c3}$, with $B = 2.25 \times 10^4$. Independently, $B = 2.25 \times 10^4$ yields an accuracy of 49.58\%, which is lower than that of $\mathcal{B}_{c3}$, standing at 50.22\%.
However, when incorporated into the set, instead of causing a decrease, it brings a further enhancement of 0.06\%, enabling $\mathcal{B}_{c4}$ to attain the highest accuracy.

Indeed, there are certain base values in this task that come quite close to the ``optimal'' accuracy. 
For instance, when we set $B$ to be $2.00 \times 10^4$, it independently yields an accuracy of 50.28\%.
Recap what we discussed in the challenges outlined in \S\ref{sec:2.3}, when we alter the task or even vary the input length, the optimal base value changes accordingly. 
In practice, enumerating bases, as we did in this experiment, becomes unfeasible.
Based on this reason and the fact that setting $B = 2.00 \times 10^4$ only outperforms the general $\mathcal{B}_{c2}$ and $\mathcal{B}_{c3}$ by a marginal 0.03\%,  the effectiveness of our approach is not undermined.

\begin{table}[t]
\vspace{2mm}
    \centering
    \resizebox{\linewidth}{15mm}{\begin{tabular}{lccc}\toprule
     & $(N,S)$ & Searched Results ($\times 10^{4}$) \\\hline
        $\mathcal{B}_{c1}$ & (7, 100) & \{1.00, 1.77, 1.78, 1.90, 2.02, 2.47, 2.48\}  \\
         $\mathcal{B}_{c2}$ & (7, 1,000) & \{1.00, 1.70, 1.80, 1.90, 2.00, 2.30, 2.50\}  \\
         $\mathcal{B}_{c3}$ & (6, 500) & \{1.00, 1.75, 1.80, 1.90, 2.00, 2.50\} \\
        $\mathcal{B}_{c4}$ & (7, 500) &  \{1.00, 1.75, 1.80, 1.90, 2.00, 2.25, 2.50\} \\\hline
        $\mathcal{B}_{A.S.1}$ & (7, -) & \{1.00, 1.30, 1.60, 1.90, 2.20, 2.50, 2.80\} \\
        $\mathcal{B}_{A.S.2}$ & (6, -) & \{1.00, 1.40, 1.80, 2.20, 2.60, 3.00\} \\\bottomrule
    \end{tabular}}
    \caption{Searched bases by variant search algorithms.}
    \label{tab:ab}
\end{table}

\begin{table}[t]
\vspace{2mm}
    \centering
 \resizebox{0.4\linewidth}{32mm}{
     \begin{tabular}{lc}\toprule
     $\mathcal{B}\ (\times 10^{4})$ & Acc. \\\hline
        \{1.00\} & 48.56 \\
        \{1.75\}& 50.10\\
        \{1.80\}& 50.01\\
        \{1.90\} & 50.00 \\
        \{2.00\} & 50.28 \\
        \{2.25\}& 49.58\\
        \{2.50\} & 49.53 \\\hline
$\mathcal{B}_{A.S.1}$& 50.19 \\
 $\mathcal{B}_{A.S.2}$ & 49.89 \\\hline
$\mathcal{B}_{c1}$ & 50.25 \\
$\mathcal{B}_{c2}$ & 50.25 \\ 
$\mathcal{B}_{c3}$  & 50.22 \\
\textbf{$\mathcal{B}_{c4}$ }  & \textbf{50.31}\\
        \bottomrule
    \end{tabular}}
    \caption{Accuracy of the $\mathcal{B}$ variations on NQ (10 documents).}
    \label{tab:ab2}
\end{table}

\section{Related Work: LLM-Based Tool-Use}
Taking advantage of large language models to use external tools is an emerging research topic~\cite{mialon2023augmented,qin2023tool}.
Researchers have explored a variety of tools, including calculators for mathematical computations~\cite{schick2023toolformer,hao2023toolkengpt}, specialized expert models~\cite{shen2023hugginggpt}, and web API calls~\cite{qin2023toolllm, patil2023gorilla}.
In these studies, LLMs interact with users by analyzing their intents and needs. 
A tool-retriever is utilized to source relevant tools, typically in the format of documents containing details like tool names, examples of use, descriptions of functions, and arguments for those functions. 
The LLM processes these documents to choose the suitable tool, inputs the necessary arguments, and relays the tool's output back to the users.
Many studies~\cite{zhou2023webarena, patil2023gorilla, qin2023toolllm} have found that large language models, including GPT-4~\cite{openai2023gpt4}, often exhibit hallucinations, such as inventing non-existent functions and arguments, or failing to adapt to changes in interactive environments. 
These highlight the critical need for enhancing these models' context awareness.
Current research heavily focuses on integrating multiple reasoning pathways to address errors caused by insufficient contextual comprehension. 
These techniques include ReAct~\cite{yao2023react}, DFS tree search~\cite{qin2023toolllm}, self-consistency~\cite{wang2023selfconsistency}, etc. 
In contrast, our approach focuses on a fundamental solution: enhancing contextual awareness.
It is both orthogonal to and stackable with those reasoning methods. 

\section{Conclusion}
In this paper, we delved into the waveform patterns observed in attention scores and found that waveform of the attention score could potentially affect the model's context awareness, particularly in relation to the position of crucial information within the context. 
We propose \name, an inference augmentation method designed to enhance the model's context awareness. 
This method combines various attention patterns, which are controlled by different RoPE bases. 
Our approach has achieved state-of-the-art performance on the current largest tool-use benchmark while showing the applicability to a wider range of RAG tasks.

\section*{Acknowledgement}
This work was supported by Alibaba Group through Alibaba Innovative Research Program.

\newpage
\bibliography{custom}
\bibliographystyle{unsrt}

\appendix
\newpage
\section{The Waveform of Attention Score Before Softmax}
\label{apx:upper-bound}

Formally, the inner-product between query vectors at position $m$ and key vectors at position $n$ within Transformer models utilizing RoPE can be formulated as:
\begin{equation}
\small
\begin{aligned}
    & q_{m}=R_{\theta, m} W_{q} x_{m},\quad
    k_{n}=R_{\theta, n} W_{k} x_{n},\\
    & q_{m} \cdot k_{n} = (R_{\theta, m} q)^{T} (R_{\theta, n} k) = \\ 
   & \text{Re}\left[\sum^{d/2-1}_{j=0} q[2j:2j+1]k^{*}[2j:2j+1] e^{i(m-n)\theta_{j}}\right] \\
    &= \sum^{d/2-1}_{j=0} (q_{2j} \cdot k_{2j} + q_{2j+1} \cdot k_{2j+1} ) \cos \left((m-n)\theta_{j}\right) \\
    & + (q_{2j} \cdot k_{2j+1} - q_{2j+1} \cdot k_{2j} ) \sin \left((m-n)\theta_{j}\right),
\end{aligned}
\label{eq:re}
\end{equation}
where $x$ is the $d$-dimensional input of the current Transformer layer, and $\theta_j = B^{-\frac{2j}{d}}$.

When $q_m$ and $k_n$ are identical, the inner-product reaches its maximum value. 
For computational simplicity, we assume them as all-one vectors to derive the waveform ($\mathcal{W}$) of the inner-product (attention score before softmax):
\begin{equation}
\begin{aligned}
    \mathcal{W} = \sum^{d/2-1}_{j=0} 2 \cos \left((m-n)\theta_{j}\right) \geq q_{m} \cdot k_{n}.
\end{aligned}
\label{eq:ub}
\end{equation}

Figure~\ref{fig:motivation}(c) illustrates the visualization of $\mathcal{W}$ with base = 10,000. 
Figure~\ref{fig:selection}(b) results from varying base with values from our searched set $\mathcal{B}_{c}$.
These figures demonstrate the horizontal axis as the relative position between $k_n$ and $q_m$. 

\newpage
\section{Locating Peaks and Troughs in an Attention Waveform}
\label{apx:calcu}

 \begin{lstlisting}[language={Python}]
d = 128 # dimension of Q or K vectors in Llama.
MAX_CONTEXT_LENGTH = 4096 # the maximum pre-trained context length.

# Calculate the waveform of attention score before softmax. 
def qmkn(base, pos_mn):
    # base: RoPE base.
    # pos_mn: relative token position for qm and kn.
    # return: the waveform of attention score between qm and kn.
    score = 0.0
    for i in range(0, d/2):
        score += 2 * np.cos((pos_mn) * np.power(base, (-2*i/d))
    return score


# Find n peak positions 
# within MAX_CONTEXT_LENGTH.
def fp(base, n,  period):
    # base: RoPE base.
    # n: expected number of searched peaks.
    # period: init approximate period.
    # return: P, a list contains peak positions.
    scores = [qmkn(pos_mn, base) for pos_mn in \ range(MAX_CONTEXT_LENGTH)]
    P = []
    start = 0
    while len(P) < n:
        p_max = np.argmax(scores[start: start \
                + period]).index + start
        P.append(p_max)
        start = p_max
        
        # The attention waveform exhibts 
        # irregualr period throughout the 
        # context, with each successive 
        # period being approximately 1.5 
        # times longer than the previous one.
        period *= 1.5 
    return P


# Find n trough positions 
# within MAX_CONTEXT_LENGTH.
def ft(base, n, period):
    # base: RoPE base.
    # n: expected number of searched troughs.
    # period: init approximate period.
    # return: T, a list contains trough positions.
    scores = [qmkn(pos_mn, base) for pos_mn in \ range(MAX_CONTEXT_LENGTH)]
    T = []
    start = 0
    while len(T) < n:
        t_min = np.argmin(scores[start: start +\ period]).index + start
        T.append(t_min)
        start = t_min
        period = 1.5 * period
    return T
\end{lstlisting}

\section{Supplement to The In-Context Retrieval Experiment}
\label{apx:kv}

There are several critical settings to make our experiments fair and convincing:

$\bullet$ \textbf{Position Anchoring Based on Last Token}: We anchor the position of each target key-value pair at its last token. 
This approach is based on the findings of~\cite{wang2023label} that the sentence semantic is gathered to the last token.

$\bullet$ \textbf{Precise Positioning of Key-Value Pairs}: To accurately place key-value pairs at specific positions within the context, we insert padding tokens after the key-value JSON data and prior to the query.

$\bullet$ \textbf{Consistent Prompt Lengths Across Rounds}: 
Since our experiments involve comparing the accuracy between two rounds, it is essential to mitigate any potential biases arising from varying context lengths.  
To achieve this, we maintain the consistency in the context length across two rounds by inserting padding tokens at the beginning of the input.

\begin{figure}[h]
    \centering
    \includegraphics[width=\linewidth]{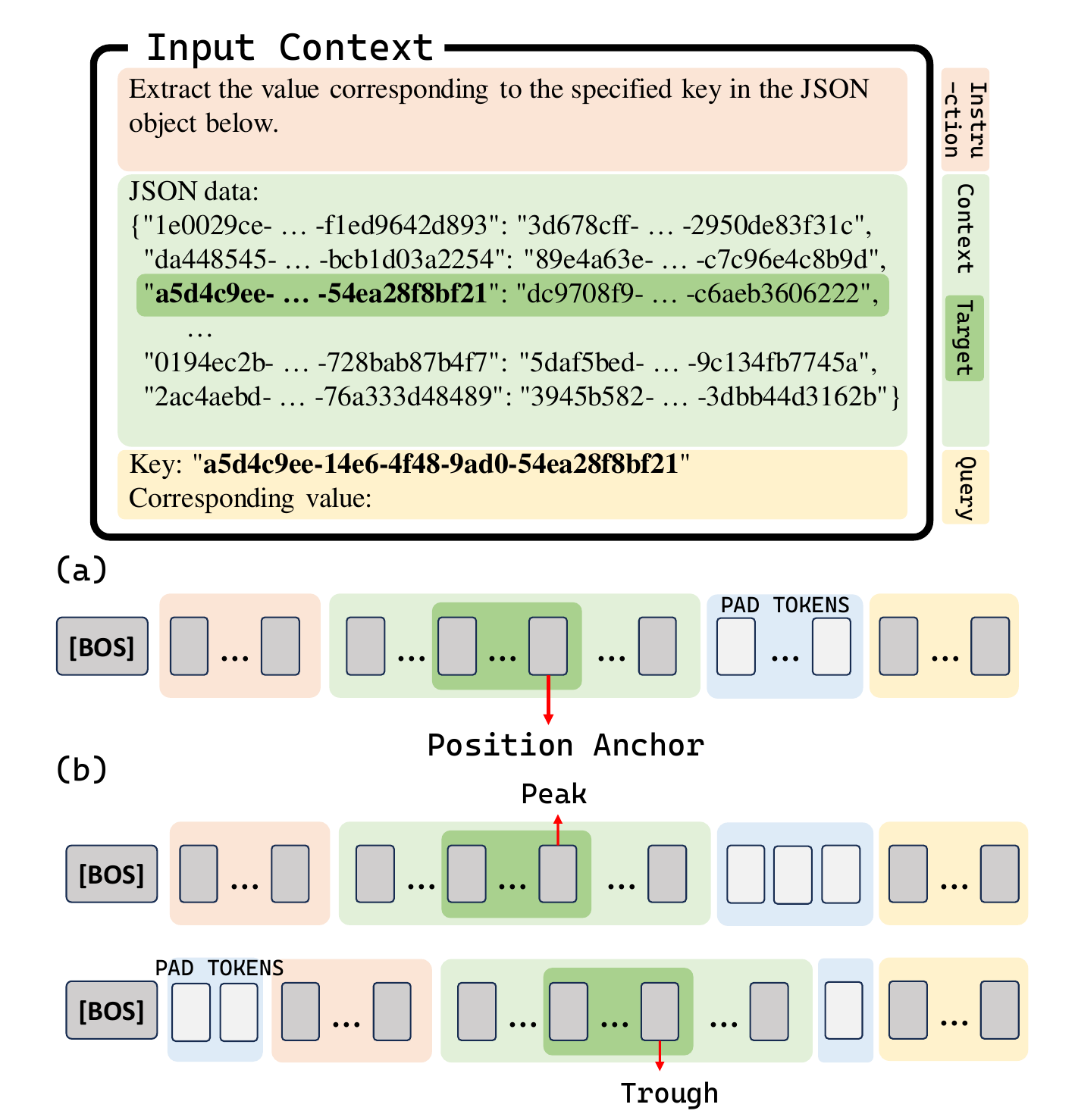}
    \caption{
    Fair and convincing experimental operations.
    (a) We apply padding prior to the "Query" to accurately locate the final token of the target key-value pair at a desired position, which corresponds to an attention peak or trough.
    (b) We use paddings to maintain consistency in prompt length across various rounds.}
    \label{fig:insert-pad}
\end{figure}

Figure~\ref{fig:insert-pad} illustrates details on above operations.

\newpage
\section{The Impact of Base Value Smaller Than That Used In Training}
\label{apx:ood-pos}
Considering Eq.\ref{eq:re}, where the position information is integrated using $d/2$ sinusoidal functions with the frequency $\theta_j = B^{-\frac{2j}{d}}$, for $j \in [0, \frac{d}{2}]$. 
If a smaller value of $B'$ is employed, compared to the pre-trained $B$, the frequency of these sinusoidal functions will be higher, resulting in a reduced period. 
In this case, given the maximum pre-trained context length, the final few tokens could correspond to positions within a period of the sinusoidal functions that are not encountered during training. 
These positions would be considered out-of-distribution for the model.
We recommend that readers interested in a more in-depth analysis of this context ``scaling law'' refer to~\cite{liu2023scaling}.

\section{Results on ToolAlpaca}
\label{apx:more-benchmarks}
\paragraph{ToolAlpaca} (Tang et al., 2023) is a framework coordinates a collection of various tools through a multi-agent simulation. 
It constructs a dataset that includes 426 unique tools across 50 categories, totaling 3,938 instances. 
The corpus is then used to fine-tune Llama, resulting in the development of two LLMs for tool-use: ToolAlpaca-7B and 13B.

\paragraph{Experiment Sets and Result} 
To assess the tool-use capabilities of language models, ToolAlpaca has developed an evaluation dataset comprised of two subsets: one encompassing 10 simulated tool APIs and the other encompassing 11 real-world tool APIs.
Each API involves various user queries which require specific functional calls and function parameters.
The evaluation relies on GPT-4 for scoring, with a primary focus on three crucial metrics:

$\bullet$ Procedure: GPT-4 assesses the model's skill in choosing the right actions, using the correct parameters, and avoid redundant steps.

$\bullet$ Response: GPT-4 verifies whether the model's output aligns with user queries.

$\bullet$ Overall: GPT-4 assesses the precision of the entire action-response cycle.

Due to the incomplete reproducibility of the open-source code of ToolAlpaca, our evaluations were limited to the simulator set.
We report the experiment results in Table~\ref{tab:al}. 
The data in the table clearly illustrate substantial enhancements our method has brought to the performance of both ToolAlpaca-7B and 13B. 
When implemented with our method, the 13B model has reached performance on par with GPT-3.5 in terms of the Overall metric. 
The experiments conducted on this benchmark demonstrate the generalizability and effectiveness of our method.

\begin{table}[t]
    \centering
    \resizebox{\linewidth}{18mm}{
\begin{tabular}{l|ccc}
\toprule
Model          & \multicolumn{1}{l}{Procedure} & \multicolumn{1}{l}{Response} & \multicolumn{1}{l}{Overall} \\ \hline
GPT-3.5       & 77.0& 85.0& 75.0                       \\ \hline
Vicuna-7B      & 19.0 &21.0 &17.0                          \\
ToolAlpaca-7B  &63.0 &69.0 &60.0                          \\
\ + \name& 69.0                             & 73.0                           &  65.0                        \\ \hline
Vicuna-13B     & 17.0 &31.0 &16.0                          \\
ToolAlpaca-13B & 70.0 & 73.0 &70.0                         \\
\ + \name & 75.0                             & 78.0                           &    74.0         \\\bottomrule
\end{tabular}}
\caption{Experimental results were obtained within the simulated tools environment using the ToolAlpaca evaluation dataset.}
\label{tab:al}
\end{table}

\end{document}